\documentclass[10pt,twocolumn,letterpaper]{article}

\usepackage{cvpr}
\usepackage{times}
\usepackage{epsfig}
\usepackage{graphicx}
\usepackage{amsmath}
\usepackage{amssymb}
\usepackage{algorithm}
\usepackage{algorithmic}
\usepackage{subfigure}
\usepackage{authblk}

\usepackage[breaklinks=true,bookmarks=false,colorlinks]{hyperref}
\cvprfinalcopy 

\def\cvprPaperID{***} 

\ifcvprfinal\fi
\begin{document}

\title{Towards Discriminability and Diversity:\\ Batch Nuclear-norm Maximization under Label Insufficient Situations}

\author{\vspace{-0.2in}Shuhao Cui$^{1,2}$ ~~ Shuhui Wang$^{1}$\thanks{Corresponding author.} ~~ Junbao Zhuo$^{1,2}$ ~~ Liang Li$^{1}$ ~~ Qingming Huang$^{1,2,3}$ ~~ Qi Tian$^{4}$\\
	$^{1}$Key Lab of Intell. Info. Process., Inst. of Comput. Tech., CAS, Beijing, China\\
	$^{2}$University of Chinese Academy of Sciences, Beijing, China\\
	$^{3}$Peng Cheng Laboratory, Shenzhen, China ~~  $^{4}$Noah's Ark Lab, Huawei Technologies \\
	{\tt\small \{cuishuhao18s, wangshuhui, liang.li\}@ict.ac.cn, junbao.zhuo@vipl.ict.ac.cn, qmhuang@ucas.ac.cn, tian.qi1@huawei.com\vspace{-0.15in}}}

\maketitle
\thispagestyle{empty}

\begin{abstract}
   The learning of the deep networks largely relies on the data with human-annotated labels. In some label insufficient situations, the performance degrades on the decision boundary with high data density. A common solution is to directly minimize the Shannon Entropy, but the side effect caused by entropy minimization, {\it i.e.}, reduction of the prediction diversity, is mostly ignored. To address this issue, we reinvestigate the structure of classification output matrix of a randomly selected data batch. We find by theoretical analysis that the prediction discriminability and diversity could be separately measured by the Frobenius-norm and rank of the batch output matrix. Besides, the nuclear-norm is an upperbound of the Frobenius-norm, and a convex approximation of the matrix rank. Accordingly, to improve both discriminability and diversity, we propose Batch Nuclear-norm Maximization (BNM) on the output matrix. BNM could boost the learning under typical label insufficient learning scenarios, such as semi-supervised learning, domain adaptation and open domain recognition. On these tasks, extensive experimental results show that BNM outperforms competitors and works well with existing well-known methods. The code is available at \url{https://github.com/cuishuhao/BNM}.
\end{abstract}

\section{Introduction}

Deep neural networks have achieved large success in most computer vision applications. Despite the success already achieved, deep models in visual learning tasks largely rely on vast amounts of labeled data, where the labeling process is both time-consuming and expensive. Without sufficient amount of labeled training data, as a common consequence, spurious predictions will be made even with a subtle departure from the training samples. Actually, in most applications, there exists large discrepancy between training data and real-world testing data. The discrepancy could lead to annoying ambiguous predictions, especially under label insufficient situations. In this paper, we focus on enhancing the model learning capability by reducing the ambiguous predictions.

\begin{figure}[t]
	\begin{center}
		\includegraphics[width=0.9\linewidth]{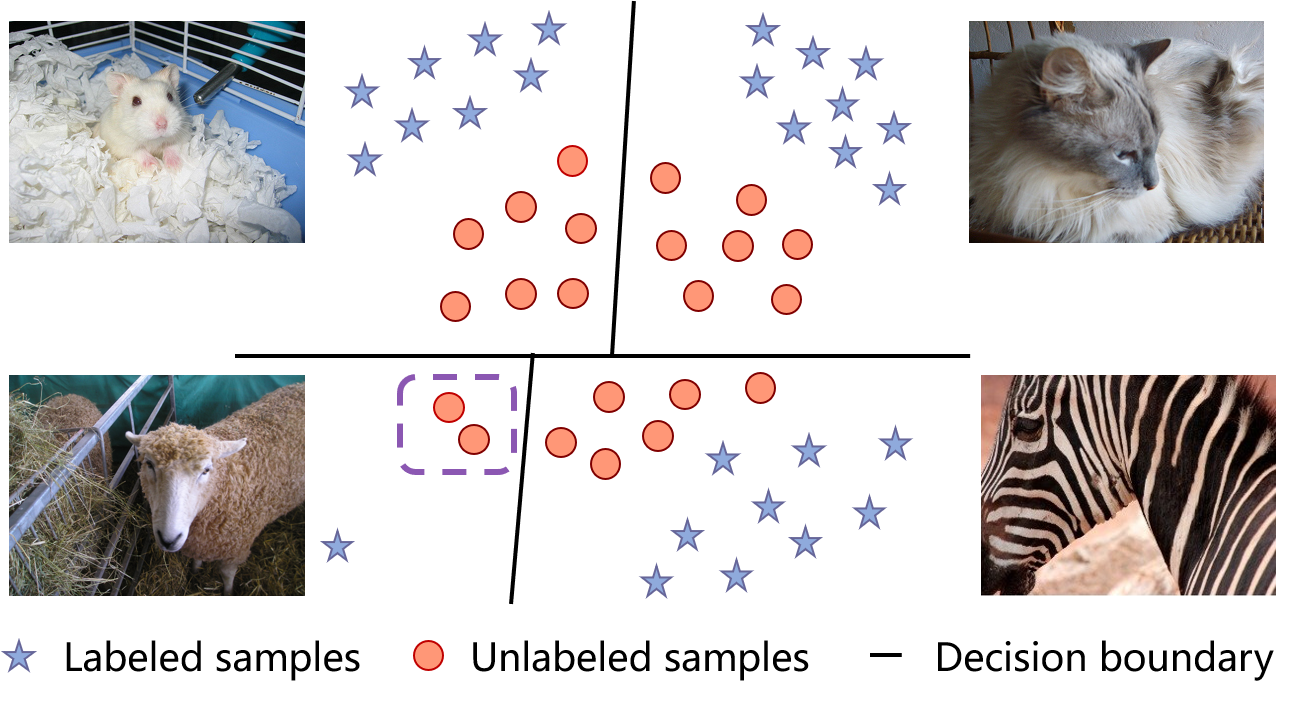}
	\end{center}
	\vspace{-2ex}
	\caption{Illustration of problem. Beyond the discriminability, we also focus on the minority category with few samples bounded in the dotted frame at the bottom left part. The influence of minority category tends to be reduced in direct entropy minimization, resulting in the degradation of the category prediction diversity.}
	\vspace{-1ex}
	\label{intro1}
\end{figure}

When deep models are applied to unlabeled examples, the prediction discriminability is always low, due to the large data density near the decision boundary. To reduce the large data density, most methods resort to classical Shannon Entropy theory~\cite{shannon1948mathematical}. In semi-supervised learning, the prediction entropy for unlabeled examples is directly minimized by \cite{grandvalet2005semi}, and further combined with Virtual Adversarial Training \cite{miyato2018virtual} for better results. In domain adaptation, the entropy minimization is also utilized in~\cite{residuallong,vu2019advent} to strengthen the discriminability on the unlabeled domain. Meanwhile, from another aspect, to encourage the prediction diversity, methods in~\cite{ubias,zhuo2019unsupervised} utilize a balance constraint to equilibrate the distribution between different categories. Methods in \cite{zou2018unsupervised,zou2019confidence} take the model predictions as pseudo-labels of the statistical distribution, to further enhance the robustness of prediction on categories with less data instances.

However, entropy-based methods suffer from the side effect of the entropy minimization, {\it i.e.}, reduction of prediction diversity. Entropy minimization pushes the examples to nearby examples far from the decision boundary. Since there are more examples in majority categories, examples are prone to be pushed into majority categories, including the examples actually belonging to minority categories. As shown in Figure~\ref{intro1}, the points in the dotted frame will be misclassified, if all the samples are classified into the other categories, resulting in the reduced prediction diversity. Towards higher prediction diversity, the balance constraint requires prior knowledge on minority categories, but it is difficult to obtain. Some methods rely on the prior knowledge estimated by pseudo-labels, to increase the prediction probability on minority categories. Nevertheless, it appears that the guidance provided by prior knowledge is usually less useful and straightforward towards more accurate prediction.

In this paper, we reinvestigate the above issues, and try to enforce the prediction discriminability and diversity of the unlabeled data. We start by looking at the structure of classification output matrix of a randomly selected data batch. 
We find by theoretical analysis that the discriminability and diversity could be measured by the Frobenius-norm and rank of the batch output matrix, respectively. The nuclear-norm of a matrix is bounded by the Frobenius-norm of the matrix. Maximizing nuclear-norm ensures large Frobenius-norm of the batch matrix, which leads to increased discriminability. The nuclear-norm of batch matrix is also a convex approximation of the matrix rank, which refers to the prediction diversity. Accordingly, we present Batch Nuclear-norm Maximization (BNM), an approach to maximize the nuclear-norm of the batch output matrix. Thus BNM could simultaneously enhance the discriminability and diversity of the prediction results.

We apply BNM to semi-supervised learning, domain adaptation and unsupervised open domain recognition to validate its effectiveness. Experiments show that our method outperforms other well-established methods on four datasets. Among these tasks, we achieve state-of-the-art results on unsupervised open domain recognition. For further validation, we observe that BNM leads to better average prediction diversity on the batch samples in the experiments. Our contribution is summarized as follows:
\begin{itemize}
	\item We theoretically prove that the discriminability and diversity of the prediction output can be measured by Frobenius-norm and rank of the batch output matrix.
	\item We propose Batch Nuclear-norm Maximization, which is a new learning paradigm that achieves better discriminability and diversity under label insufficent learning situations.
	\item We achieve promising performance on semi-supervised learning, domain adaptation and open domain recognition.
\end{itemize}

\section{Related Work}

In this paper, we analyze the label insufficient situations on three typical tasks, {\it i.e.}, semi-supervised learning~\cite{kingma2014semi,odena2016semi,kipf2016semi,wangs3mkl}, domain adaptation~\cite{gretton2012kernel,long2015learning} and unsupervised open domain recognition. Semi-supervised learning~\cite{tarvainen2017mean,miyato2018virtual} leverages the unlabeled examples to improve the robustness of model. Domain adaptation~\cite{hoffman2017cycada,saito2018maximum,long2018conditional,zhuo2017deep} reduces the domain discrepancy between labeled and unlabeled examples. Unsupervised Open Domain Recognition~\cite{zhuo2019unsupervised} considers a more realistic circumstance than domain adaptation, where some categories in unlabeled domain are unseen in labeled domain.

Among the tasks, they all face the problem of the rich data distribution near the decision boundary. To reduce the ambiguous predictions, most methods resort to the Shannon Entropy~\cite{shannon1948mathematical} to model uncertainty. In semi-supervised learning, reducing the entropy of the classification responses for unlabeled domain is adopted in \cite{grandvalet2005semi}. Meanwhile, the entropy minimization is further applied to Virtual Adversarial Training in \cite{miyato2018virtual} and implicitly modeled by pseudo-labels~\cite{berthelot2019mixmatch}. In domain adaptation, entropy minimization is utilized in~\cite{residuallong,vu2019advent} to obtain reliable prediction on unlabeled examples. The entropy minimization is further modified into maximum squares loss in \cite{chen2019domain} to lower the influence of easy-to-transfer samples.

To maintain prediction diversity on the minority categories, a direct thought is resorting to imbalanced learning~\cite{he2009learning}. Existing imbalanced learning methods such as \cite{ubias,zhuo2019unsupervised} enforce the ratio of predictions on minority categories to be appropriately higher. However, they demand prior knowledge on the category distribution. Without the prior knowledge, the predictions are taken as pseudo-labels in \cite{zou2019confidence,zou2018unsupervised} to approximate the category distribution, while our method is performed in a data driven manner that is free from any form of prior knowledge. From another aspect of increasing diversity, Determinantal Point Processes (DPPs)~\cite{kulesza2012determinantal} act probabilistically to capture the balance between quality and diversity within sets, but suffer from the large computation time. 

In this paper, we analyze the problem from the perspective of matrix analysis~\cite{candes2009exact,cai2010singular}, which has already been widely applied to many computer vision tasks. As a popular perspective, matrix completion is based on the assumption that the noisy data brings extra components to the matrix. To reduce the influence of the extra components, minimizing nuclear-norm of the matrix has been applied to image denoising~\cite{gu2014weighted}, image restoration~\cite{dong2012nonlocal} and many other tasks. In comparison to the above methods, we aim to explore the extra information in the matrix, thus the nuclear-norm of the matrix is maximized towards more prediction diversity. Recently, BSP~\cite{BSP_ICML_19} penalizes the largest singular values of the batch feature matrix to boost the feature discriminability for domain adaptation, while we analyze the batch classification response matrix to increase both the prediction discriminability and diversity.

\section{Method}
\subsection{Measuring Discriminability with $F$-norm}
\label{fnorm}
In the training process of a deep neural network, we start by looking at the prediction outputs on a data batch with $B$ randomly selected unlabeled samples. Denote the number of categories as $C$, and we represent the batch prediction output matrix as $\mathrm{A} \in \mathbb{R}^{B \times C}$, which satisfies:
\begin{equation}
\begin{split}{\mathrm
	\sum_{j=1}^C\mathrm{A}_{i,j}=1 \quad &\forall i \in 1...B \\
	\mathrm{A}_{i,j} \ge 0 \quad &\forall i \in 1...B, j \in 1...C,
}\end{split}
\label{matrix}
\end{equation}
where deep methods could achieve well-performed response matrix $\mathrm{A}$ by training with sufficient number of labeled samples. However, in label insufficient situations, the discrepancy between labeled and unlabeled data might result in the high-density regions of the marginal data distribution near the task-specific decision boundary. Since the ambiguous samples are easy to be misclassified, we focus on optimizing prediction results of unlabeled samples by increasing the discriminability.

Actually, higher discriminability means less uncertainty in the prediction. To measure the uncertainty, most methods resort to Shannon Entropy~\cite{shannon1948mathematical}, which is always denoted as entropy for simplicity. The entropy could be calculated as follows:
\begin{equation}
\begin{split}{
	{H}(\mathrm{A}) = -\frac{1}{B}{\sum_{i=1}^B \sum_{j=1}^C \mathrm{A}_{i,j}log(\mathrm{A}_{i,j}) }.
}\end{split}
\label{trace}
\end{equation}
Same as~\cite{grandvalet2005semi,miyato2018virtual,residuallong,vu2019advent}, we could directly minimize $H(\mathrm{A})$, towards lessened uncertainty and more discriminability.
When $H(\mathrm{A})$ reaches the minimum, only one entry is $1$ and other $C-1$ entries are $0$ in each row of $\mathrm{A}_i$, {\it i.e.}, $\mathrm{A}_{i,j} \in \{0,1\} \quad \forall i \in 1...B, j \in 1...C$. The minimum exactly satisfies the highest prediction discriminability of $\mathrm{A}$, where each prediction $\mathrm{A}_i$ is fully determined.

Other functions could improve the prediction discriminability, by push $\mathrm{A}$ to the same minimum with $H(\mathrm{A})$. We choose to calculate Frobenius-norm ($F$-norm) $\left\| \mathrm{A} \right\|_F$, as follows:
\begin{equation}
\begin{split}{
	\left\| \mathrm{A} \right\|_F = \sqrt{\sum_{i=1}^B \sum_{j=1}^C |\mathrm{A}_{i,j}|^2 }.
}\end{split}
\label{trace}
\end{equation}
We prove that $H(\mathrm{A})$ and $\left\|\mathrm{A}\right\|_F$ have strict opposite monotonicity and the minimum of $H(\mathrm{A})$ and the maximum of $\left\| \mathrm{A} \right\|_F$ could be achieved at the same value in Supplementary. Particularly, according to inequality of arithmetic and geometric means, the upper-bound of $\left\| \mathrm{A} \right\|_F$ could be calculated as:
\begin{equation}
\begin{split}{
	\left\| \mathrm{A} \right\|_F &\leq \sqrt{\sum_{i=1}^B (\sum_{j=1}^C \mathrm{A}_{i,j}) \cdot (\sum_{j=1}^C \mathrm{A}_{i,j})}\\
	&= \sqrt{\sum_{i=1}^B 1 \cdot 1 }\quad= \sqrt{B}.
}\end{split}
\label{tracemax}
\end{equation}
The upper-bound of $\left\| \mathrm{A} \right\|_F$ could be achieved in the same $\mathrm{A}$ with the minimum of $H(\mathrm{A})$. Thus prediction discriminability could also be enhanced by maximizing $\left\| \mathrm{A} \right\|_F$.

\subsection{Measuring Diversity with Matrix Rank}
\label{rank}

It is normal in randomly selected batch of $B$ examples that some categories dominate the samples, while other categories contain less or even no samples. In this case, a model trained with entropy minimization or $F$-norm maximization tends to classify samples near the decision boundary to the majority categories. The continuous convergence to the majority categories reduces the prediction diversity, which is harmful to the overall prediction accuracy. To improve the accuracy, different from \cite{ubias,zhuo2019unsupervised,zou2018unsupervised,zou2019confidence}, we aim to maintain the prediction diversity by analyzing the batch output matrix $\mathrm{A}$ to model the diversity.

To model prediction diversity, we start by looking at the fixed batch of $B$ unlabeled samples in matrix $\mathrm{A}$. The number of categories in the predictions is expected to be a constant on average. If this constant becomes larger, the prediction approach could obtain more diversity. Thus the prediction diversity could be measured by the number of predicted categories in the batch output matrix $\mathrm{A}$.

We further analyze the number of categories and the predicted vectors in $\mathrm{A}$. Two randomly selected prediction vectors, {\it i.e.}, $\mathrm{A}_i$ and $\mathrm{A}_k$, could be linearly independent when $\mathrm{A}_i$ and $\mathrm{A}_k$ belong to different categories. When $\mathrm{A}_i$ and $\mathrm{A}_k$ belong to the same category and $\left\| \mathrm{A} \right\|_F$ is near $\sqrt{B}$, the differences between $\mathrm{A}_i$ and $\mathrm{A}_k$ are tiny. Then $\mathrm{A}_i$ and $\mathrm{A}_k$ could be approximately regarded as linearly dependent. The largest number of linear independent vectors is called the matrix rank. Thus $rank(\mathrm{A})$ could be an approximation on the number of predicted categories in $\mathrm{A}$, if $\left\| \mathrm{A} \right\|_F$ is near the upper-bound $\sqrt{B}$.

Based on the above analysis, the prediction diversity could be approximately represented by $rank(\mathrm{A})$, when $\left\| \mathrm{A} \right\|_F$ is near $\sqrt{B}$. Accordingly, we could maximize $rank(\mathrm{A})$ to maintain prediction diversity. Apparently, the maximum value of $rank(\mathrm{A})$ is $\min(B, C)$. When $B \geq C$, the maximum value is $C$, which firmly guarantees that the prediction diversity on this batch achieves the maximum. However, when $B < C$, the maximum value is less than $C$, it still enforces that the predictions on the batch samples should be as diverse as possible, though there is no guarantee that all the categories will be assigned to at least one sample. Therefore, maximization of $rank(\mathrm{A})$ could ensure the diversity in any case.

\subsection{Batch Nuclear-norm Maximization}
\label{theory}

For a normal matrix, the calculation of the matrix rank is an NP-hard non-convex problem, and we could not directly restrain the rank of matrix $\mathrm{A}$. Theorem in~\cite{fazel2002matrix} shows that when $\left\|\mathrm{A}\right\|_F \leq 1$, the convex envelope of $rank(\mathrm{A})$ is the nuclear-norm $\left\| \mathrm{A} \right\|_{\star}$. In our situation, different from above theorem, we have $\left\|\mathrm{A}\right\|_F \leq \sqrt{B}$ as shown in Eqn.~\ref{tracemax}. Thus the convex envelope of $rank(\mathrm{A})$ becomes $ \left\| \mathrm{A} \right\|_{\star} / \sqrt{B}$, which is also proportional to $\left\| \mathrm{A} \right\|_{\star}$. Meanwhile, $rank(\mathrm{A})$ could approximately represent the diversity, when $\left\| \mathrm{A} \right\|_F$ is near the upper-bound, as described in Sec.~\ref{rank}. Therefore, if $\left\| \mathrm{A} \right\|_F$ is near $\sqrt{B}$, the prediction diversity could be approximately represented by $\left\| \mathrm{A} \right\|_{\star}$. Also, maximizing $\left\| \mathrm{A} \right\|_{\star}$ could ensure higher prediction diversity.

\begin{figure}[t]
	\begin{center}
		\includegraphics[width=0.98\linewidth]{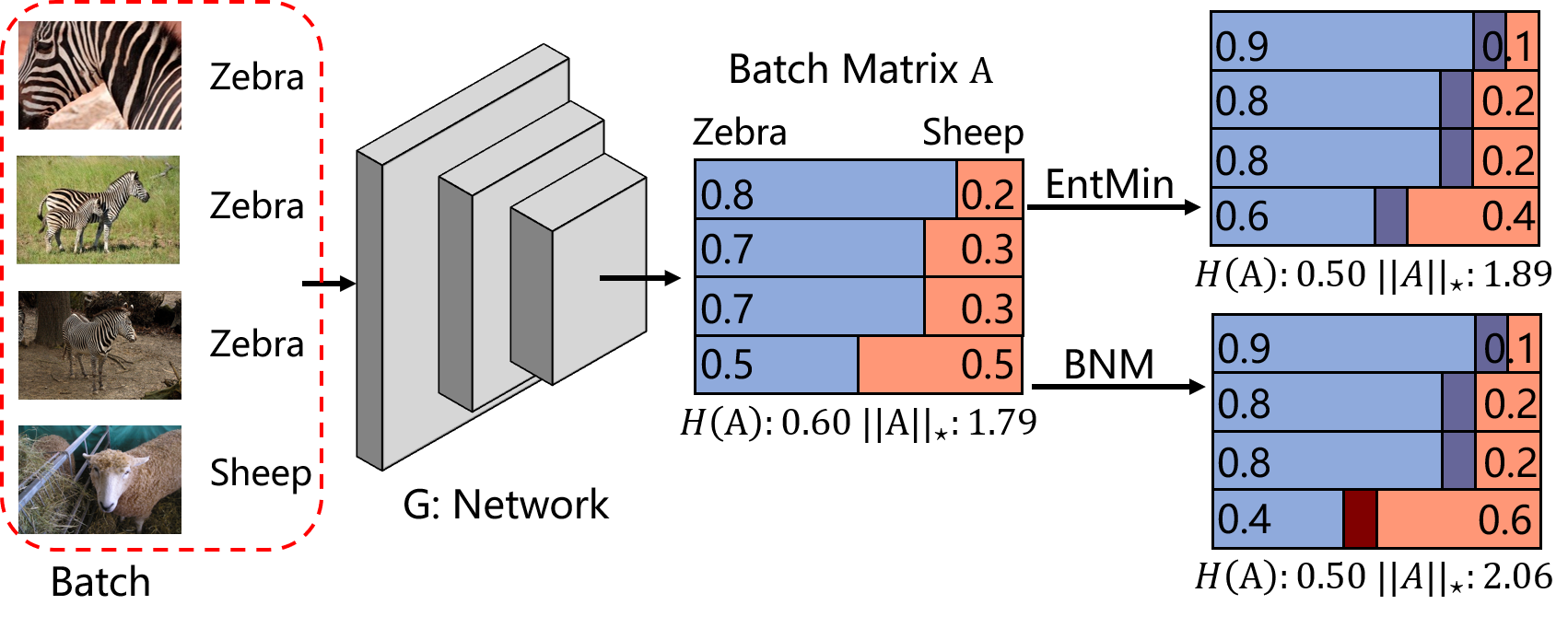}
	\end{center}
	\caption{Illustration of comparison between the effect of BNM and Entropy Minimization (EntMin) in a toy example with two categories and batch size 4. The dark region means the increase of the variable, {\it i.e.}, the dark blue (red) represents the increase of blue (red) variable. $H(\mathrm{A})$ represents the entropy value and $\left\| \mathrm{A} \right\|_{\star}$ represent the value of nuclear-norm.}
	\label{softmatching}
\end{figure}

In \cite{fazel2002matrix,recht2010guaranteed,srebro2005maximum}, the relationship of the range between $\left\| \mathrm{A} \right\|_{\star}$ and $\left\| \mathrm{A} \right\|_F$ could be expressed as follows:
\begin{equation}
\begin{split}{
	\frac{1}{\sqrt{D}}\left\| \mathrm{A} \right\|_{\star} \leq \left\| \mathrm{A} \right\|_F \leq \left\| \mathrm{A} \right\|_{\star} \leq \sqrt{D} \cdot \left\| \mathrm{A} \right\|_F
}\end{split}
\label{lep}
\end{equation}
where $D=\min(B,C)$. This shows that $\left\| \mathrm{A} \right\|_{\star}$ and $\left\| \mathrm{A} \right\|_F$ could bound each other. 
Therefore, $\left\| \mathrm{A} \right\|_F$ tends to be larger, if $\left\| \mathrm{A} \right\|_{\star}$ becomes larger. 
Since maximizing $\left\| \mathrm{A} \right\|_F$ could improve the discriminability described in Sec.~\ref{fnorm}, maximizing $\left\| \mathrm{A} \right\|_{\star}$ also contributes to the improvement on prediction discriminability.

Due to the relationship between $\left\| \mathrm{A} \right\|_{\star}$ and $\left\| \mathrm{A} \right\|_F$, and the fact that upper-bound of $\left\| \mathrm{A} \right\|_F$ is $\sqrt{B}$ in Eqn.~\ref{tracemax}, we could calculate the maximum of $\left\| \mathrm{A} \right\|_{\star}$ as follows:
\begin{equation}
\begin{split}{
	\left\| \mathrm{A} \right\|_{\star}	\leq \sqrt{D} \cdot \left\| \mathrm{A} \right\|_F \leq \sqrt{D \cdot B},
}\end{split}
\label{lep}
\end{equation}
where we could find that the influence factor of $\left\| \mathrm{A} \right\|_{\star}$ could be separated into two parts, respectively corresponding to the two inequality conditions in the equation. The first inequality corresponds to the diversity, and the second corresponds to the discriminability. When the diversity is larger, the rank of $\mathrm{A}$ tends to be larger and $\left\| \mathrm{A} \right\|_{\star}$ tends to increase. Similarly, when the discriminability becomes larger, $\left\| \mathrm{A} \right\|_F$ tends to increase and $\left\| \mathrm{A} \right\|_{\star}$ tends to be larger.

Based on the above findings, maximizing $\left\| \mathrm{A} \right\|_{\star}$ could lead to the improvement on both the prediction discriminability and diversity. Thus to improve discriminability and diversity, we propose Batch Nuclear-norm Maximization by maximizing the nuclear-norm of the batch matrix $\mathrm{A}$, where $\mathrm{A}$ represents the classification responses on a batch of $B$ randomly selected samples. For better comprehension of the effect of BNM, we build two toy examples, one explaining BNM in the maximum case, and another in ordinary situations. 

In the first example, we assume $B$ and $C$ are $2$. In this case, $\mathrm{A}$ could be expressed as:
\begin{equation}
\mathrm{A}=\left[
\begin{array}{ccc} 
x &    1-x   \\ 
y &    1-y  \\ 
\end{array}
\right],
\end{equation}
where $x$ and $y$ are variables. Thus the entropy, $F$-norm and nuclear-norm could be calculated as:
\begin{equation}
\begin{split}{
	H(\mathrm{A})=&-x\log(x)-(1-x)\log(1-x)-y\log(y)\\
	&-(1-y)\log(1-y)\\
	\left\| \mathrm{A} \right\|_F =& \sqrt{x^2+(1-x)^2+y^2+(1-y)^2}\\
	\left\| \mathrm{A} \right\|_{\star} =& \sqrt{x^2+(1-x)^2+y^2+(1-y)^2+2|y-x|},
}\end{split}
\end{equation}
where the calculation of $\left\| \mathrm{A} \right\|_{\star}$ is described in Supplementary. For entropy and $F$-norm, there is no constraint limiting the relationship between $x$ and $y$, thus entropy and $F$-norm could reach the optimal solution when:
\begin{equation}
\mathrm{A}=\left[
\begin{array}{ccc} 
1 &    0   \\ 
1 &    0  \\ 
\end{array}
\right],
\left[
\begin{array}{ccc} 
0 &    1   \\ 
1 &    0  \\ 
\end{array}
\right],
\left[
\begin{array}{ccc} 
1 &    0   \\ 
0 &    1  \\ 
\end{array}
\right],
\left[
\begin{array}{ccc} 
0 &    1   \\ 
0 &    1  \\ 
\end{array}
\right].
\end{equation}
But $\left\| \mathrm{A} \right\|_{\star}$ would reach the optimal solution when: 
\begin{equation}
\mathrm{A}=
\left[
\begin{array}{ccc} 
0 &    1   \\ 
1 &    0  \\ 
\end{array}
\right],
\left[
\begin{array}{ccc} 
1 &    0   \\ 
0 &    1  \\ 
\end{array}
\right].
\end{equation}
where $\left\| \mathrm{A} \right\|_{\star}$ tries to maintain diversity by maximizing the prediction divergence among the data batch in this example.

The second example is shown in Figure~\ref{softmatching}, where we assume that there are only two categories, {\it i.e.}, zebra and sheep. In this data batch with three zebras and a goat, the zebra category is the majority category. The matrices after Entropy Minimizataion and BNM could obtain the same value on entropy as $0.50$. But for the matrices, the value of nuclear-norm tends to be larger when the sheep is correctly classified. Direct entropy minimization tends to classify the batch examples into the majority category zebra. In comparison, BNM could maintain the prediction possibility of minority classes, and the image of sheep tends to be assigned with the infrequent but correct label.

Since BNM is computed by matrix operations, there remain concerns on the computational complexity. To obtain the nuclear-norm, we calculate all the singular values in the matrix $\mathrm{A}$. The singular value decomposition computing on the matrix $\mathrm{A} \in \mathbb{R}^{B \times C}$ costs $O(\min(B^2C, BC^2))$ time. Since the batch size $B$ is always small, the overall computational budget of $\left\| \mathrm{A} \right\|_{\star}$ is almost negligible in the training of deep networks.

\subsection{Application}

We apply BNM to three typical label insufficient situations, including semi-supervised learning, unsupervised domain adaptation and unsupervised open domain recognition. In the tasks, we are given labeled domain $\mathcal{D}_L$ and unlabeled domain $\mathcal{D}_U$. There are $N_L$ labeled examples $\mathcal{D}_L=\{(x_i^L,y_i^L)_{i=1}^{N_L} \}$ in $C$ categories and $N_U$ unlabeled examples $\mathcal{D}_U=\{(x_i^U)_{i=1}^{N_U}\}$. In $\mathcal{D}_L$, the labels are denoted as $y_i^L =[y_{i1}^L,y_{i2}^L,...,y_{iC}^L]\in \mathbb{R}^{C}$, where $y_{ij}^L$ equals to $1$ if $x_i^L$ belongs to the $j$th category otherwise $0$. 

In these tasks, the classification responses are obtained by the deep network $G$, {\it i.e.}, $\mathrm{A}_i=G(x_i)$. The classification network consists of a feature extraction network, a classifier and a softmax layer. With randomly sampled batch size $B_L$ examples $\{X^L,Y^L\}$ on the labeled domain, the classification loss on $\mathcal{D}_L$ could be calculated as:
\begin{equation}
\mathcal{L}_{cls} =\frac{1}{B_L} \left \| Y^L{log(G(X^L))}\right \|_1,
\label{cls} \end{equation}
where the classification loss could provide initial parameters for further optimization.

For learning on the unlabeled domain $\mathcal{D}_U$, on all the three tasks, we apply the method BNM introduced in Sec.~\ref{theory} to the classification response matrix. With randomly sampled batch size $B_U$ examples $\{X^U\}$, the classification response matrix on $D_U$ could be denoted as $G(X^U)$. And the loss function of BNM can be formulated as:
\begin{equation}
\mathcal{L}_{bnm}= -\frac{1}{B_U}\left \|G(X^U)\right \|_{\star} ,
\label{opt}
\end{equation}
where the neural network $G$ is shared between both $D_L$ and $D_U$. Minimizing $\mathcal{L}_{bnm}$ could reduce the data density near the decision boundary without losing diversity, which is more effective than typical entropy minimization. Meanwhile, the gradient of nuclear-norm could be calculated according to \cite{papadopoulo2000estimating}, thus $\mathcal{L}_{bnm}$ could be applied to the training process of gradient-based deep networks.

To train the network, we simultaneously optimize classification loss and BNM loss, {\it i.e.}, $\mathcal{L}_{cls}$ and $\mathcal{L}_{bnm}$ could be simultaneously optimized and combined with the parameter $\lambda$ as follows:
\begin{equation}
\mathcal{L}_{all}=\frac{1}{B_L} \left \| Y^L{log(G(X^L))}\right \|_1 -\frac{\lambda}{B_U} \left \|G(X^U)\right \|_{\star}.
\label{opt}
\end{equation}
By enforcing diversity, the key insight of BNM may be sacrificing a certain level of the prediction hit-rate on majority categories, to enhance the prediction hit-rate on minority categories. The samples belonging to the majority classes might be misclassified as minority classes, to increase the diversity. But the classification loss on the labeled training data would penalize the wrongly encouraged diversity in a batch, since classification loss is simultaneously minimized. Asymptotically, the network tends to produce more diverse prediction given that the samples can be correctly predicted. As a consequence, BNM is particularly useful to avoid prediction degradation for learning in label insufficient situations on datasets with both balanced and imbalanced category distributions.

\section{Experiments}
We apply our method to semi-supervised learning, unsupervised domain adaptation, and unsupervised open domain recognition. The experiments of the three tasks are done on CIFAR-100~\cite{krizhevsky2009learning}, Office-31 \cite{saenko2010adapting}, Office-Home \cite{venkateswara2017deep} and I2AwA~\cite{zhuo2019unsupervised}. The results with the notion of ${method}^*$ are reproduced by us in the same environment with our methods, while other results are directly reported from the original papers. We also denote the direct entropy minimization, batch Frobenius-norm maximization as EntMin, BFM in our experiments. When applied to the existing methods, we denote Batch Nuclear-norm Maximization or entropy minimization as +BNM or +EntMin.
\subsection{Semi-supervised Learning}
CIFAR-100~\cite{krizhevsky2009learning} is a standard benchmark dataset for semi-supervised learning. We evaluate our method of BNM on CIFAR-100 with 5000 and 10000 labeled examples respectively. We utilize the ResNet~\cite{he2016deep} model, the same backbone with~\cite{oliver2018realistic}. The batch size is fixed to $64$ in our experiments. The experiments are implemented with Tensorflow~\cite{abadi2016tensorflow}. We create $4$ splits for each and report the mean and variance across the accuracy on different splits. 

The results are shown in Table~\ref{Cifar100}. In semi-supervised learning (SSL), direct entropy minimization could improve the performance, while BNM outperforms entropy minimization. The improvement of BNM applied on a simple pretrained ResNet is moderate compared to other state-of-the-art well-designed SSL methods. However, working with other SSL methods such as VAT~\cite{miyato2018virtual}, BNM demonstrates more significant improvement, which is comparable to methods with more complicated mechanism such as ML+CCN+VAT~\cite{wu2019mutual}. Thus BNM is more suitable for cooperation with existing SSL methods, and performs better than entropy minimization in all cases.

\subsection{Domain Adaptation}

Office-31 \cite{saenko2010adapting} and Office-Home \cite{venkateswara2017deep} are standard benchmarks for domain adaptation. Office-31 contains 4,652 images in 31 categories, and consists of three domains: Amazon (A), Webcam (W), and DSLR (D). We evaluate the methods across the three domains, resulting in six transfer tasks. Office-Home is a relative challenging dataset with 15,500 images in 65 categories. It has four significantly different domains: Artistic images (Ar), Clip Art (Cl), Product images (Pr), and Real-World images (Rw). There are 12 challenging transfer tasks among four domains in total. 
\begin{table}[t]
	\caption{Accuracy(\%) on the CIFAR-100 dataset for semi-supervised learning methods.}
	\vspace{-4ex}
	\begin{center}
		\scalebox{1.0}{
			\begin{tabular}{ccc}
				\hline             Method        &5000        &10000           \\
				\hline\hline
				Temporal Ensembling~\cite{laine2016temporal}    &- &61.35$\pm$0.51 \\
				SNTG+$\Pi$-model~\cite{luo2018smooth}&-&62.03$\pm$0.29\\
				ML+CCN+VAT~\cite{wu2019mutual} &56.58$\pm$0.31 &\textbf{64.72}$\pm$0.23\\
				\hline
				ResNet \cite{he2016deep}&39.73$\pm$0.33&49.55$\pm$0.28\\
				EntMin&40.92$\pm$0.18&50.36$\pm$0.20\\
				BNM                        &41.59$\pm$0.27&51.07$\pm$0.24\\
				\hline
				VAT*~\cite{miyato2018virtual}&56.63$\pm$0.18&63.62$\pm$0.18\\
				VAT+EntMin &56.97$\pm$0.21 &64.48$\pm$0.22\\
				VAT+BNM             &\textbf{57.43}$\pm$0.24	&{64.61}$\pm$0.15  \\
				
				\hline
				
			\end{tabular}
		}
	\end{center}
	\label{Cifar100}
	\vspace{-3ex}
\end{table}
\begin{table}[t]
	\begin{center}
		
		
		\caption{Accuracies (\%) on Office-31 for ResNet50-based unsupervised domain adaptation methods.}
		\vspace{-10pt}
		\label{tableoffice}
		\scalebox{0.93}{
			\setlength{\tabcolsep}{0.3mm}{ 
				\begin{tabular}{cccccccc}
					\hline
					Method & A$\rightarrow$D & A$\rightarrow$W & D$\rightarrow$W & W$\rightarrow$D & D$\rightarrow$A & W$\rightarrow$A & Avg \\
					\hline\hline
					ResNet-50 \cite{he2016deep} & 68.9 & 68.4 & 96.7 & 99.3 & 62.5 & 60.7 & 76.1 \\
					GFK \cite{gong2012geodesic} & 74.5 & 72.8 & 95.0 & 98.2 & 63.4 & 61.0 & 77.5 \\
					DAN \cite{long2015learning} & 78.6 & 80.5 & 97.1 & 99.6 & 63.6 & 62.8 & 80.4 \\
					DANN \cite{ganin2016domain} & 79.7 & 82.0 & 96.9 & 99.1 & 68.2 & 67.4 & 82.2 \\
					ADDA \cite{tzeng2017adversarial} & 77.8 & 86.2 & 96.2 & 98.4 & 69.5 & 68.9 & 82.9 \\
					MaxSquare \cite{chen2019domain}&90.0 &92.4 &\textbf{99.1} &\textbf{100.0}  &68.1 &64.2 &85.6\\
					Simnet \cite{pinheiro2018unsupervised} & 85.3 & 88.6 & 98.2 & 99.7 & {73.4} & 71.8 & 86.2 \\				
					GTA \cite{sankaranarayanan2018generate} & 87.7 & 89.5 & 97.9 & 99.8 & 72.8 & 71.4 & 86.5 \\
					MCD \cite{saito2018maximum} & {92.2} & {88.6} & {98.5} & \textbf{100.0} & {69.5} & {69.7} & {86.5} \\	
					CBST~\cite{zou2018unsupervised}  &86.5  &87.8  &98.5 &\textbf{100.0}  &70.9  &71.2  &85.8 \\
					CRST~\cite{zou2019confidence}   &88.7  &89.4  &98.9 &\textbf{100.0} &70.9  &72.6 &86.8\\

					\hline
					EntMin  &86.0	&87.9	&98.4	&\textbf{100.0} 	&67.0		&63.7&83.8\\
					BFM &87.7	&86.9	&98.5	&\textbf{100.0}	&67.6	&63.0	&84.0 \\
					
					BNM  & {90.3} & {91.5} & {98.5} & \textbf{100.0} & {70.9}  & {71.6} & {87.1} \\
					\hline
					CDAN \cite{long2018conditional} & \textbf{92.9} & \textbf{93.1} & {98.6} & \textbf{100.0} & {71.0} & {69.3} & {87.5} \\
					CDAN+EntMin &92.0	&91.2 &98.7	&\textbf{100.0}	&70.7		&71.0	&87.3\\
					CDAN+BNM  & \textbf{92.9} & {92.8} & {98.8} & \textbf{100.0} & \textbf{73.5} & \textbf{73.8} & \textbf{88.6} \\
					\hline
			\end{tabular}}
			
		}
	\end{center}
	\vspace{-20pt}
\end{table}

\begin{table*}[htbp]
	\begin{center}
		\vspace{-5pt}
		
		\caption{Accuracies (\%) on Office-Home for ResNet50-based unsupervised domain adaptation methods.}
		\label{tableofficehome}
		\addtolength{\tabcolsep}{-5.5pt}
		\scalebox{0.93}{
			\resizebox{\textwidth}{!}{%
				\begin{tabular}{cccccccccccccc}
					\hline
					Method& Ar$\rightarrow$Cl & Ar$\rightarrow$Pr & Ar$\rightarrow$Rw & Cl$\rightarrow$Ar & Cl$\rightarrow$Pr & Cl$\rightarrow$Rw & Pr$\rightarrow$Ar & Pr$\rightarrow$Cl & Pr$\rightarrow$Rw & Rw$\rightarrow$Ar & Rw$\rightarrow$Cl & Rw$\rightarrow$Pr & Avg \\
					\hline\hline
					
					ResNet-50 \cite{he2016deep} & 34.9 & 50.0 & 58.0 & 37.4 & 41.9 & 46.2 & 38.5 & 31.2 & 60.4 & 53.9 & 41.2 & 59.9 & 46.1 \\
					DAN \cite{long2015learning} & 43.6 & 57.0 & 67.9 & 45.8 & 56.5 & 60.4 & 44.0 & 43.6 & 67.7 & 63.1 & 51.5 & 74.3 & 56.3 \\
					DANN \cite{ganin2016domain} & 45.6 & 59.3 & 70.1 & 47.0 & 58.5 & 60.9 & 46.1 & 43.7 & 68.5 & 63.2 & 51.8 & 76.8 & 57.6 \\
					MCD \cite{saito2018maximum} &48.9	&68.3	&74.6	&61.3	&67.6	&68.8	&57	&47.1	&75.1	&69.1	&52.2	&79.6	&64.1 \\
					
					SAFN~\cite{Xu_2019_ICCV} &52.0&	71.7&76.3&64.2&69.9&71.9&63.7&51.4&	77.1&70.9&57.1&81.5&67.3\\
					Symnets \cite{zhang2019domain} & 47.7 & 72.9 & 78.5 & \textbf{64.2} & 71.3 & 74.2 & \textbf{64.2}  & 48.8 & 79.5 & \textbf{74.5} & 52.6 & {82.7} & 67.6 \\
					MDD \cite{zhang2019bridging} & 54.9 & 73.7 & 77.8 & 60.0 & 71.4 & 71.8 & 61.2 & 53.6 & 78.1 & 72.5 & \textbf{60.2} & 82.3 & 68.1 \\
					
					\hline
					
					EntMin  &43.2	&68.4	&78.4	&61.4	&69.9	&71.4	&58.5	&44.2	&78.2	&71.1	&47.6	&81.8	&64.5 \\
					BFM &43.3 	&69.1 	&78.0 	&61.3 	&67.4 	&70.9 	&57.8 	&44.1 	&78.9 	&72.1 	&50.1 	&81.0 	&64.5 \\
					BNM &{52.3}	&\textbf{73.9}	&\textbf{80.0}	&63.3	&{72.9}	&\textbf{74.9}	&61.7	&{49.5}	&{79.7}	&70.5	&{53.6}	&82.2	&67.9\\	
					
					\hline
					CDAN \cite{long2018conditional} & 50.7 & 70.6 & 76.0 & 57.6 & 70.0 & 70.0 & 57.4 & 50.9 & 77.3 & 70.9 & 56.7 & {81.6} & 65.8 \\
					CDAN+EntMin &54.1	&72.4	&78.3	&61.8	&71.8	&73.0	&62.0	&52.3	&79.7	&72.0	&57.0	&83.2	&68.1 \\

					CDAN+BNM &\textbf{56.2}	&{73.7}	&{79.0} 	&63.1 	&\textbf{73.6}	&{74.0} &62.4	&\textbf{54.8}	&\textbf{80.7}	&72.4	&58.9	&\textbf{83.5}	&\textbf{69.4}\\
					
					\hline
		\end{tabular}}}
		\vspace{-18pt}
	\end{center}
\end{table*}

We adopt ResNet-50 \cite{he2016deep} pre-trained on ImageNet \cite{deng2009imagenet} as our backbone. The batch size is fixed to $36$ in our experiments. The experiments are implemented with PyTorch~\cite{torch}. BNM loss is directly combined with classification loss with the parameter $\lambda$ fixed to $1$. When BNM is combined with existing methods, the parameter $\lambda$ is fixed to $0.1$. For each method, we run four random experiments and report the average accuracy. 

\begin{figure}[t]
	\begin{center}
		\begin{minipage}{.23\textwidth}
			\subfigure[Ar $\rightarrow$ Cl]{	
				\includegraphics[width=0.9\textwidth]{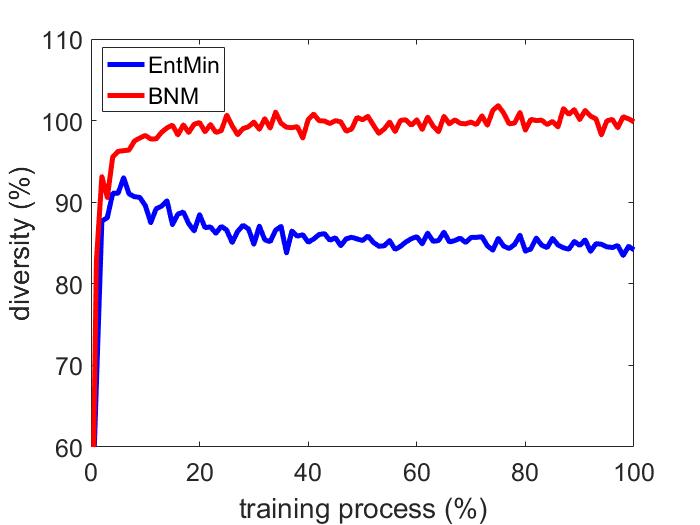}
				\label{home1}}
		\end{minipage}
		\begin{minipage}{.23\textwidth}
			\subfigure[Ar $\rightarrow$ Pl]{
				\includegraphics[width=0.9\textwidth]{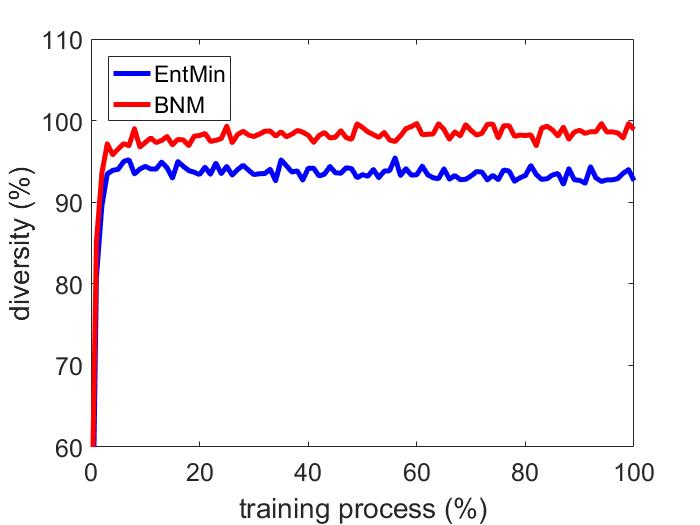}
				\label{home2}}
		\end{minipage}
		\caption{Diversity ratio on Office-Home for domain adaptation, calculated as the predicted diversity divided by the ground truth diversity. The predicted (ground truth) diversity is measured by the average number of predicted (ground truth) categories in randomly sampled batches.}
		\label{officenum}
	\end{center}
	\vspace{-3ex}
\end{figure}
\begin{table}
	\caption{Accuracies (\%) on I2AwA for ResNet50-based unsupervised open domain recognition methods.}
	\begin{center}
		\label{I2AwA}
		\vspace{-3pt}
		\scalebox{1.0}{
			\begin{tabular}{ccccc}
				\hline               Method      &Known        &Unknown         &All      &Avg    \\
				\hline\hline
				zGCN~\cite{zgcn}    &77.2         &21.0            &65.0    &49.1    \\
				dGCN~\cite{adgpm}            &78.2        &11.6            &64    &44.9     \\
				adGCN~\cite{adgpm}           &77.3         &15.0           &64.1   &  46.2    \\
				bGCN~\cite{ubias}            &84.6         &28.0           &72.6     &56.3   \\	
				pmd-bGCN~\cite{pmd}          &84.7         &27.1           &72.5    &55.9   \\	
				UODTN~\cite{zhuo2019unsupervised}                 &84.7 &31.7   &73.5&58.2\\
				
				Balance*~\cite{zhuo2019unsupervised}  	&85.9 &22.3	&72.4 &54.1\\	
				
				\hline
				EntMin 	&87.5&7.2	&70.5	&47.4\\
				BFM	&87.7 &9.2	&71.1	&48.4\\				
				
				BNM                        &\textbf{88.3}&\textbf{39.7}   &\textbf{78.0}&\textbf{64.0}\\
				\hline
			\end{tabular}
		}
	\end{center}
	\label{I2AwA}
	\vspace{-3ex}
\end{table}

The results on Office-31 and Office-Home are shown in Table~\ref{tableoffice} and~\ref{tableofficehome}. On both Office-31 and Office-Home, as we expected, BFM obtains similar results with EntMin, while BNM achieves substantial improvement on average over other entropy-based methods. Surprisingly, BNM obtains superior results compared with popular alignment-based comparison methods. The results show that Batch Nuclear-norm Maximization is effective for domain adaptation, especially on the difficult tasks where the baseline accuracy is relatively low. Besides, we add BNM to existing CDAN~\cite{long2018conditional}, denoted as CDAN+BNM. The results of CDAN+BNM outperforms CDAN and CDAN+EntMin by a large margin, which shows that BNM could cooperate well with other methods. In summary, BNM could not only be regarded as a basic simple method for domain adaptation, but also an effective module contributing to existing methods.

To validate that BNM could maintain the diversity in domain adaptation compared with entropy minimization, we show the diversity ratio in Office-Home on tasks of Ar $\rightarrow$ Cl and Ar $\rightarrow$ Pr in Figure~\ref{officenum}. The diversity is measured by the mean matrix rank, {i.e.}, mean number of predicted categories in randomly sampled batch. Thus the diversity ratio is measured by the mean predicted category number dividing the mean ground-truth category number. As shown in Figure~\ref{home1}, the diversity ratio of BNM is larger than that of the EntMin by a large margin in Ar $\rightarrow$ Cl. This phenomenon is normal since the rich samples near the decision boundary are mainly classified into the majority categories, reducing the diversity in the batch examples. As shown in Figure~\ref{home2}, the diversity of BNM is still larger in Ar $\rightarrow$ Pl. But the differences between EntMin and BNM in diversity ratio are shorter than that in Ar $\rightarrow$ Cl. This results from fewer samples near the decision boundary in Ar $\rightarrow$ Pl since Ar $\rightarrow$ Pl is easier, {\it i.e.}, the basic accuracy of Ar $\rightarrow$ Pl is higher. Thus BNM is more effective in difficult tasks with rich data near the decision boundary.
\subsection{Unsupervised Open Domain Recognition}
\label{exp}

\begin{figure*}
	\begin{center}
		\begin{minipage}{.98\textwidth}
			\begin{minipage}{.24\textwidth}
				\subfigure[All Category Accuracy]{	
					\includegraphics[width=0.9\textwidth]{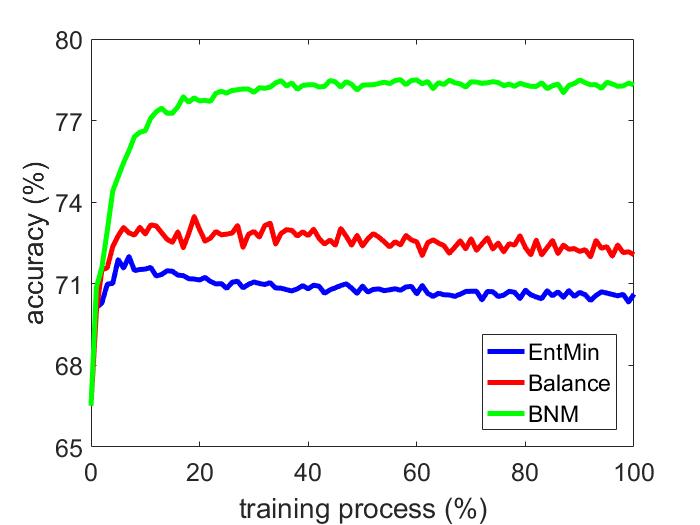}
					\label{compare1}}
			\end{minipage}
			\begin{minipage}{.24\textwidth}
				\subfigure[Known Category Accuracy]{
					\includegraphics[width=0.9\textwidth]{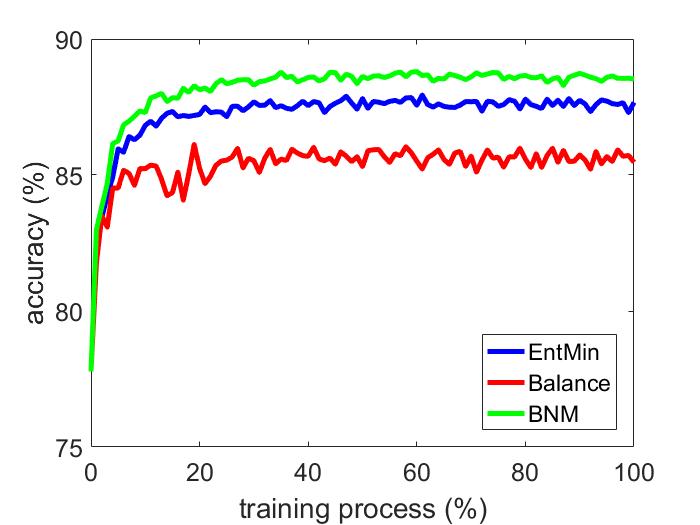}
					\label{compare0}}
			\end{minipage}
			\begin{minipage}{.24\textwidth}
				\subfigure[Unknown Category Accuracy]{
					\includegraphics[width=0.9\textwidth]{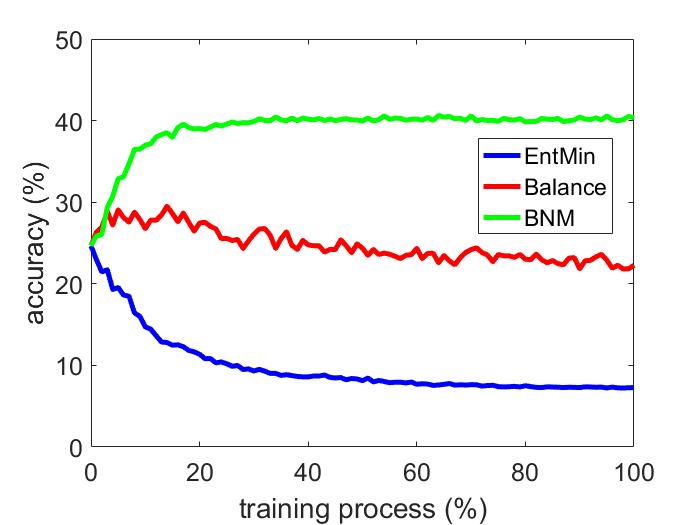}
					\label{compare3}}
			\end{minipage}
			\begin{minipage}{.24\textwidth}
				\subfigure[Unknown Category Ratio]{
					\includegraphics[width=0.9\textwidth]{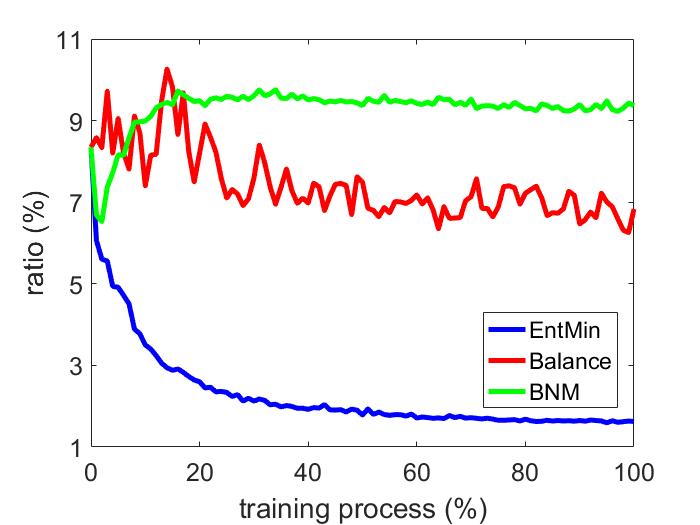}
					\label{compare2}}
			\end{minipage}

			
			\caption{Statistics for Entropy, Balance and Nuclear-norm in the whole training process.}
			\label{compare}
		\end{minipage}
	\end{center}
	\vspace{-2ex}
\end{figure*}

We evaluate our BNM method on I2AwA for unsupervised open domain recognition~\cite{zhuo2019unsupervised}. In I2AwA, the labeled domain consists of 2,970 images belonging to 40 known categories, via selecting images from ImageNet and Google image search engine. The unlabeled domain of I2AwA is AwA2~\cite{aw2} which contains a total of 37,322 images. The images are totally classified into 50 categories, with the same 40 known categories as labeled domain, and the remaining 10 classes as unknown categories. 

To obtain a reliable initial classification model on unknown categories, we construct the same knowledge graph for I2AwA with UODTN~\cite{zhuo2019unsupervised}. The graph structure is built according to the popular methods, Graph Convolutional Networks (GCN) \cite{kipf2016semi,zgcn}. The graph nodes include all categories in the unlabeled domain and also their children and ancestors in WordNet~\cite{wordnet}. To obtain the features of the nodes, we choose the word vectors of all categories extracted via the GloVe text model~\cite{GloVe} trained on Wikipedia. We use ResNet-50~\cite{he2016deep} pretrained on ImageNet~\cite{deng2009imagenet} as our backbone, where the parameters of the last fully connected layer could be initialized by the parameters of GCN in the same categories.

For fair comparison, we perform in the same environment as UODTN~\cite{zhuo2019unsupervised}. The experiments are implemented with Pytorch~\cite{torch}. We fix the batch size to $48$ for both the labeled and unlabeled domain. We apply BNM on the classfication outputs on the total 50 categories and minimize classification loss on the known 40 categories in labeled domain and BNM loss on all the 50 categories in the unlabeled domain to train the network. We report the results of known categories, unknown categories, all categories on unlabeled domain and the average of known and unknown category accuracy. For each method, we run four random experiments and report the average classification accuracy. 

\begin{table}
	\caption{Parameter Sensitivity on the I2AwA dataset for ResNet50-based unsupervised open domain recognition methods.}
	\begin{center}
		\label{I2AwA}
		\scalebox{1.0}{
			\begin{tabular}{ccccc}
				\hline                Method     &Known        &Unknown         &All      &AVG    \\
				\hline	\hline
				BNM ($\lambda=1$)                       &{88.0}&{39.4}   &{77.7}&{63.7}\\
				BNM ($\lambda=1.5$)                       &{88.1}&\textbf{39.7}   &{77.9}&{63.9}\\
				BNM ($\lambda=2$)                       &\textbf{88.3}&\textbf{39.7}   &\textbf{78.0}&\textbf{64.0}\\
				BNM ($\lambda=3$)                       &{87.7}&{39.5}   &{77.5}&{63.6}\\
				BNM ($\lambda=4$)                       &{87.4}&{38.6}   &{77.1}&{63.0}\\
				\hline
			\end{tabular}
		}
	\end{center}
	\label{stable}
	\vspace{-3ex}
\end{table}
The results are shown in Table~\ref{I2AwA}, we achieve remarkable improvement on I2AwA. We achieve 11.4\% improvement on the known categories over the baseline zGCN, and BNM surprisingly improves by 19.0\% on the unknown categories over zGCN. From the overall range of the dataset, we achieve 13.3\% improvement on the whole dataset and we achieve an average improvement of 15.2\% improvement over zGCN. Besides, BNM outperforms the state-of-the-art UODTN~\cite{zhuo2019unsupervised} by 4.8\%. This shows that the simple BNM is effective enough for unsupervised open domain recognition, which outperforms the combination of complex functions in UODTN~\cite{zhuo2019unsupervised}. We also show the parameter sensitivity experiments in Table~\ref{stable}. The results show BNM is relatively stable under different parameters, and we set $\lambda$ fixed to $2$ for further comparison.

We also compare the training process of EntMin, Balance and BNM loss functions in Figure~\ref{compare}. The prediction results on all categories, known categories and unknown categories are separately shown in Figure~\ref{compare1},~\ref{compare0} and~\ref{compare3}. BNM outperforms others on All accuracy, Known category accuracy and Unknown category accuracy in the whole training process. To explore the intrinsic effect of BNM on unknown categories, we show the unknown category ratio, which is the ratio of predicting the samples in the labeled domain of I2AwA into unknown categories in Figure~\ref{compare2}. 
Obviously, Entmin reduces the unknown category ratio by a large margin, which greatly damages the prediction diversity and accuracy on unknown categories. Though the unknown category ratio in BNM is reduced at first, it gradually raises along the training process, and it appears to be even higher than initial ratio after training. This means BNM could protect the diversity by ensuring ratio of prediction on minority categories. Though the Balance constraint could also protect ratio of prediction on minority categories, results of Balance loss seems not quite stable. Besides, the accuracy of Balance loss is much lower than BNM due to the lack of discriminability. The experimental phenomenon has steadily proved the effectiveness of BNM towards both discriminability and diversity.

\subsection{Discussion}

The chosen tasks are typical label insufficient situations to show the mechanism of BNM. Among the tasks and datasets, there are differences in two aspects, {\it i.e.}, the domain discrepancy and category balance. There exists large domain discrepancy in tasks of domain adaptation and unsupervised open domain recognition, while no domain discrepancy is assumed in semi-supervised learning. From the view of category balance, the categories are balanced in datasets of semi-supervised learning, {\it i.e.}, CIFAR-100. In datasets of domain adaptation, {\it i.e.}, Office-31 and Office-Home, the categories are imbalanced. While unsupervised open domain recognition is a learning task with extremely imbalanced category distributions, where some categories are even unseen in the labeled domain. In datasets of I2AwA, 10 categories are unknown categories, which hold a remarkable percentage of the total 50 categories.

As shown in the experiments, BNM could cooperate well with existing methods in semi-supervised learning. For domain adaptation, BNM could outperform most existing methods using losses such as adversarial loss. While in unsupervised open domain recognition, method with only the BNM loss and classification loss could even achieve state-of-the-art results. We could see the progressive progress and fitness of BNM to the tasks, from semi-supervised learning to unsupervised open domain recognition. Considering the differences between the tasks, we could obtain two conclusions on the applicability of BNM. The first is that BNM could work well in label insufficient situations. The other is that BNM outperforms entropy minimization significantly, especially when there exists rich domain discrepancy and imbalanced category distribution.

\section{Conclusion}
    The discriminability and diversity could be separately represented by the Frobenius-norm and rank of the batch output matrix. Nuclear-norm is the upperbound of Frobenius-norm, also a convex approximation of matrix rank. Accordingly, we propose BNM method which maximizes the batch nuclear-norm to ensure higher prediction discriminability and diversity. Experiments show our method is suitable for the classification tasks under scenarios of semi-supervised learning, domain adaptation and open domain recognition. We will explore the effect of BNM on other settings and tasks in the future.

\vspace{4pt}
\textbf{Acknowledgement}. This work was supported in part by the National Key R\&D Program of China under Grant 2018AAA0102003, in part by National Natural Science Foundation of China: 61672497, 61620106009, 61836002, 61931008, 61771457, 61732007 and U1636214, and in part by Key Research Program of Frontier Sciences, CAS: QYZDJ-SSW-SYS013. 
\newpage
{\small
\bibliographystyle{ieee_fullname}
\bibliography{egbib}
}

\end{document}